\documentclass[letterpaper]{article} 
\usepackage{aaai25}  
\usepackage{caption}
\usepackage{subcaption}
\usepackage{times}  
\usepackage{helvet}  
\usepackage{courier}  
\usepackage[hyphens]{url}  
\usepackage{graphicx} 
\usepackage{natbib}  
\usepackage{caption} 
\usepackage{algorithm}
\usepackage{algorithmic}
\usepackage{graphicx} 
\usepackage{amsmath} 
\usepackage{amssymb} 
\usepackage{amsthm} 
\usepackage{xcolor} 
\usepackage{braket} 
\usepackage{parskip} 
\usepackage{comment} 
\usepackage{pdfpages} 
\usepackage{svg} 
\usepackage{tabularx}

\usepackage{array}
\usepackage{ragged2e}
\newcolumntype{L}[1]{>{\RaggedRight\hspace{0pt}}p{#1}}

\usepackage{newfloat}
\usepackage{listings}
\DeclareCaptionStyle{ruled}{labelfont=normalfont,labelsep=colon,strut=off} 
\floatstyle{ruled}
\newfloat{listing}{tb}{lst}{}
\floatname{listing}{Listing}
\newcommand{\RL}{RL}

\newcommand{\PV}{PV}

\newcommand{\states}{S}

\newcommand{\initState}{s_{0}}
\newcommand{\actions}{A}
\newcommand{\rewardFunc}{R}
\newcommand{\transitionFunc}{T}
\newcommand{\discountFactor}{\gamma}

\usepackage{comment}
\newcount\Comments
\usepackage{color}
\definecolor{darkgreen}{rgb}{0,0.7,0}
\definecolor{gray}{rgb}{0.95,0.95,0.95}
\definecolor{lilach}{rgb}{0.95,0.5,0.95}
\definecolor{purple}{rgb}{1,0,1}
\definecolor{teal}{rgb}{0.2,0.95,0.95}
\definecolor{orange}{rgb}{1,0.5,0}
\newcommand{\kibitz}[2]{\ifnum\Comments=1{\color{#1}{#2}}\fi}
\newcommand{\sk}[1]{\kibitz{purple}{[sk: #1]}}

\title{Comparing Traditional and Reinforcement-Learning Methods for Energy Storage Control}
\author {
    Elinor Ginzburg\textsuperscript{\rm 1},
    Itay Segev\textsuperscript{\rm 2},
    Yoash Levron\textsuperscript{\rm 1,3},
    Sarah Keren\textsuperscript{\rm 2,3}    
}
\affiliations{
    \textsuperscript{\rm 1} The Viterbi Faculty of Electrical and Computer Engineering, Technion---Israel Institute of Technology\\

    \textsuperscript{\rm 2} The Taub Faculty of Computer Science
Technion --- Israel Institute of Technology
    \\
  \textsuperscript{\rm 3} The Grand Technion Energy Program
    \\

}

\usepackage{bibentry}

\begin{document}

\maketitle

\begin{abstract}
We aim to better understand the tradeoffs between traditional and reinforcement learning (\RL) approaches for energy storage management. More specifically, we wish to better understand the performance loss incurred when using a generative \RL~policy instead of using a traditional approach to find optimal control policies for specific instances.  Our comparison is based on a simplified micro-grid model, that includes a load component, a photovoltaic source, and a storage device. Based on this model, we examine three use cases of increasing complexity: ideal storage with convex cost functions, lossy storage devices, and lossy storage devices with convex transmission losses. With the aim of promoting the principled use \RL~based methods in this challenging
and important domain, we provide a detailed formulation of each use case and a detailed description of the optimization challenges. We then compare the performance of traditional and
\RL~methods, discuss settings in which it is beneficial to use each method, and suggest avenues for future investigation. 
\end{abstract}

\section{Introduction}

Energy storage devices are vital in modern power systems especially given the increasing penetration of renewable energy sources. Power production of renewable sources such as solar and wind cannot be controlled, thus making their behaviour intermittent and unreliable. Thus, to effectively incorporate these technologies into the existing grid, storage devices must be used. In this manner, surplus energy from renewable sources during low-demand hours can be stored, and released during peak times, ensuring the stability and efficiency of the modern grid operation as well as reducing carbon emission \cite{renewable-1,renewable-2}. Moreover, the surge in electric vehicle production further amplifies the need for efficient storage control to manage the charging process intelligently to minimize strain on the grid and optimize consumption patterns \cite{ev-1,ev-2}.

\begin{figure}[t]
    \centering
    \includegraphics[width=\linewidth]{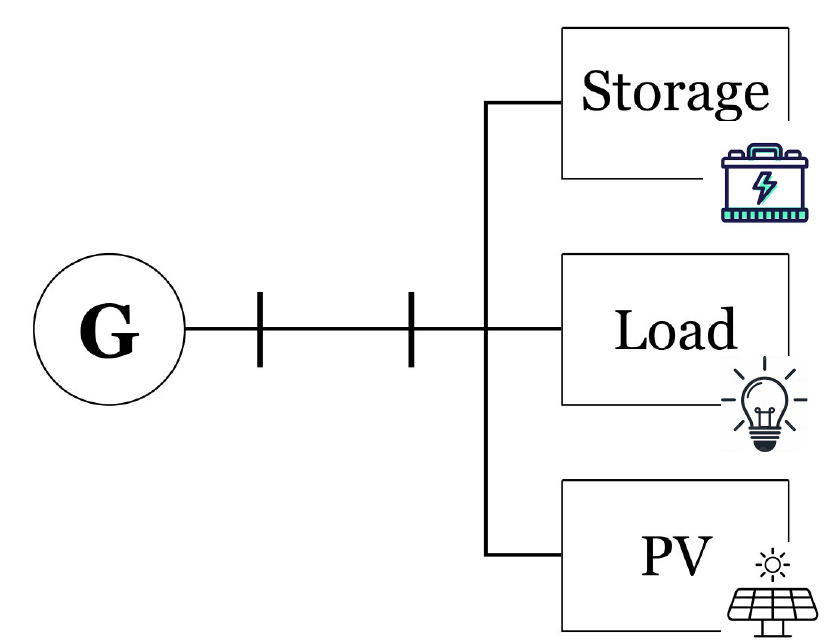}
    \caption{A simplified microgrid, comprising of a generator, a photovoltaic source (PV), a load, and a storage device.}
    \label{fig:micro-grid}
\end{figure}

Consider the simplified micro-grid represented in Figure \ref{fig:micro-grid}, with a load (consumption) component, a PV with generation capabilities, a storage device, and a connection to a generator, which may represent local fuel-based generation or the electrical grid from which power can be purchased. A storage control policy will account for the predicated load and set the current state of charge (SOC) of the battery to minimize the amount of electricity acquired from the generator. 

Traditional methods for storage control typically seek an optimal value for power to charge or discharge at every time step over a fixed horizon. Thus, the number of decision variables increases rapidly and the problem quickly becomes high-dimensional. Moreover, the optimization problems are often non-convex either because the objective function is non-convex or because the decision variables are defined over a non-convex set. As a result, gradient-based methods are often inefficient since they tend to converge to local minima. Another main limitation lies in that traditional methods heavily rely on a model of the environment and are not designed to adapt to changes in the dynamics of the environment.  Finally, traditional methods are designed to find an optimal solution for a given instance and have no generalization capabilities. Thus, each system (e.g., household) needs to be optimized separately. This may be sufficient if the objective is to optimize a single micro-grid but is insufficient if the objective is to find a general policy that will yield good performance for different micro-grids.


To address this challenge, various machine learning methods, and reinforcement learning (\RL) methods in particular have been introduced. Their strength lies in their ability to use a data-driven approach to produce high accuracy and efficient solutions, despite the high dimensional feature space. A major advantage of such approaches is that full knowledge of the system's dynamics, its input signal, and its statistics is often not required. Instead, deep-learning-based \RL~methods are designed to learn from data collected from the system and improve performance by exploring unknown aspects of the system. 
A key weakness of these approaches is that they often require a vast amount of interaction data to learn effectively. This high sample complexity makes training expensive and time-consuming, especially in real-world environments, like the one considered here, where data collection is costly or slow.


In light of this emerging trend in this crucial application, we aim to better understand the tradeoffs between the traditional and \RL~approaches to storage management. More specifically, we wish to better understand the performance loss incurred when the physical model of the storage component is not known, and a data-driven approach is used instead. Our comparison is based on the simplified micro-grid represented in Figure \ref{fig:micro-grid}, with a load component (representing consumption), a PV with generation capabilities, a storage device, and a connection to a generator, e.g., the grid. 
We examine three use cases of increasing complexity: ideal storage with convex cost functions, lossy storage devices, and lossy storage devices with non-convex cost functions.

A key characteristic of most existing approaches to storage control is that they have been developed by energy systems experts who use off-the-shelf \RL~methods which may not be the most suitable. This is mostly because energy systems are highly complex and require extensive domain knowledge that is not accessible to general AI researchers and because creating effective collaboration between researchers in these very different fields is challenging. 
Thus, in light of the emerging trend of using \RL~in energy systems and the potential societal and environmental benefits of optimizing storage control, our objective is to facilitate this bridge by formulating the storage optimization in a way that is comprehensible to AI researchers and by better understanding the effectiveness of using \RL~in this critical application. 
Our examination is based on three variations of increasing complexity of the setting depicted in Figure \ref{fig:micro-grid}.


\section{Background and Related Work}
There are several aspects to maximizing the efficiency, reliability, and safety of methods for storage control. These include monitoring the charge and discharge policy, assessment of the overall performance of the storage device over time, temperature control, and more. Our focus here is on the former challenge and on comparing traditional approaches to novel \RL~methods for finding policies that minimize the expected overall cost of power acquired from the generator.

The problem of managing grid-connected storage device is well-known and has with various approaches based on traditional control methods \cite{jiang2014a}, 
including shortest-path methods \cite{Levron2010Optimal}, Pontryagin's minimum principle \cite{Chowdhury2021Optimal}, Dynamic Programming (DP) \cite{Zargari2023An}, and Model-Predictive Control (MPC) \cite{Zhang2020MPC}. Such methods are designed to optimize the storage policy over a fixed time frame or \emph{episode}, given a model of the load, storage, and generation dynamics. While the complexity of these solutions is typically polynomial in the state-space size, 
however this size is exponential in the number of environment features and granularity of 
discretization. In addition, these methods heavily rely on prior knowledge of the system dynamics and input signals. For instance, classic use of Pontryagin's minimum principle or dynamic programming requires full knowledge of the physical model, including the dynamics of the storage device, and sometimes of the transmission or distribution dynamics. In many cases, they also require full knowledge of the load (consumption) signal. Such requirements are often not realistic, either since exact knowledge of the physical model is not available, or since the input signals cannot be predicted accurately. While stochastic optimization versions of these algorithms, such as stochastic dynamic-programming or model-predictive control \cite{Zhang2020MPC}, can model uncertainty with regards to the model of the environment, they are not designed to deal with high-dimensional space or to generalize to unseen instances. 

In contrast to traditional approaches which aim to optimize the policy for a single episode, the \RL~approaches we consider here aim to find a generalized policy over a set of episodes sampled from an underlying and unknown distribution. 
Thus, while traditional methods based on search and DP plan over a given and known time horizon $T$ (e.g., one day or one week), require a way to predict the outcomes of all possible actions, and operate until convergence 
 before returning an optimal solution, RL methods are \emph{anytime} approaches that can work with a partial model of the environment. such methods collect experiences during execution and improve performance as more experience is gained. 

Due to the growing need for generative methods that account for large populations of \emph{prosumers}, producer-consumer agents, and the inability of traditional methods to deal with this need, we are witnessing an emerging trend of \RL~methods for storage control. \RL~algorithms vary greatly: some are \emph{model-based} in that they use the acquired experience to learn the model of environment dynamics which they use to extract a policy. In contrast, \emph{model-free} approaches directly optimize policies or value functions. Nevertheless, \RL~methods are data-driven and require less prior knowledge but more data in comparison to traditional methods.

Reinforcement learning (RL) %
deals with the problem of learning policies for sequential decision-making 
in environments in which the dynamics are not fully known \cite{sutton2018reinforcement}. A common assumption is that the environment can be modeled as a Markov Decision Process (MDP), 
typically defined as a tuple $\left<\states, \initState, \actions, \rewardFunc, \transitionFunc, \discountFactor \right>$, where $\states$ is a finite set of states typically described via a set of random variables $X = {X_1, \dots, X_n}$, , $\initState \in \states$ is an initial state,  
    $\actions$ is a finite set of actions, 
    $\rewardFunc:\states \times \actions \times \states  \rightarrow \mathbb{R}$ is a Markovian and stationary reward function that specifies the reward $r(s, a, s')$ that an agent gains from transitioning from state $s$ to $s'$ by the
execution of action $a$, 
    $\mathcal{P:S}\times \mathcal{A} \rightarrow \mathbb{P}[\mathcal{S}]$ is a transition function denoting a probability distribution $p(s,a,s')$ over next states $s'$ when action $a$ is executed at state $s$, and
    $\gamma \in [0, 1]$ is a discount factor.  The objective is to find a policy that maps states to actions and maximizes the return, representing the expected total discounted reward.

Various approaches have been developed for using \RL~for storage control with the state space of the underlying MDP including the storage devices’s state of charge (SOC) and forecasted electricity prices and a reward function that balances arbitrage profit against battery degradation costs. 
For example, \citet{cao2020deep} offer a deep \RL~framework that incorporates a lithium-ion battery degradation model and employs a model-free approach with a noisy network architecture that introduces randomness directly into the network's weights. A hybrid model combining convolutional neural networks and long short-term memory networks is used for predicting electricity prices, optimizing both spatial and temporal data patterns. This approach enhances the profitability of energy storage arbitrage while strategically managing battery health for long-term operational efficiency. Another example is \cite{bui2019double}, where a Double Deep Q-Learning (DDQN) approach is used to optimize the operation of a community battery. The system considers both grid-connected and islanded modes, dynamically adjusting battery usage based on real-time market prices, internal power demands. and variable renewable energy outputs. We focus on providing a comprehensive analysis of different \RL~approaches for controlling the storage of a basic micro-grid.

Perhaps closest to our work is the comparison between traditional and \RL~by \citet{lee2020comparative} which addresses the fuel economy of hybrid electric vehicles. 
Based on simulation results, the authors show that the \RL-based strategies can obtain global optimality that is achieved by stochastic dynamic programming. 
Our objective is to examine the performance loss that may be incurred by using a policy based on \RL methods as opposed to using traditional methods for computing the policy for each new instance. 


\section{Case Studies}
We aim to optimize the policy of a storage device over a time horizon $T$. The variables at each time step are the power values per time unit $P(t)$ measured in watts.
Accordingly, our optimization considers energy, measured in watt-hours, that is the accumulated power over time which is derived from the power functions by the following relation $E(t) = \int_0^T P(\tau) \mathrm{d}\tau$. 
We let $E_g(t), E_L(t), \text{ and } E_s(t)$ denote the generated, consumed, and stored energy respectively. 

We compare classical and \RL~methods for storage control. Importantly, the classical and \RL~approaches solve different problems. The classical methods find an optimal path between two constraints in a deterministic search space, while the \RL~methods find a policy that generalizes over different load patterns and environmental conditions. Thus, optimal control methods such as Dijkstra are guaranteed to minimize the total generation cost for the specified load values. In contrast, the \RL~methods adopt a generative approach that may not optimize the policy for any specific load pattern but instead optimizes the expected value over the different examined patterns.

To examine the loss incurred for using \RL~methods as opposed to using classical methods, we use three use-cases of increasing complexity. In all settings, we consider a micro-grid comprising a grid-connected storage device, a renewable energy source (here a PV panel), and a generator, as depicted in Figure \ref{fig:micro-grid}. 

The power injected from the grid is $P_g: \mathbb{R}_{\geq 0} \to \mathbb{R}$. The  \PV~is modeled by the power it provides at each time step $t$, denoted by a piece-wise continuous semi-positive function $P_{pv}: \mathbb{R}_{\geq 0} \to \mathbb{R}$.
The load represents a device or set of devices for which the aggregated load is characterized by its active power consumption. Formally, the load power demand is a continuous non-negative function $P_a: \mathbb{R}_{\geq 0} \to \mathbb{R}$ over a finite time interval $[0,T]$ for some arbitrary and known $T$.

To satisfy the demand at each time step $t$ the load can consume power generated at time $t$ by the generator or the PV or power that was previously stored. The variable $P_s(t)$ represents the power charged or discharged from the storage at time $t$ and depends on the current demand from the load, denoted $P_L(t)$ and the power $P_g(t)$ acquired from the generator.
We assume that the net power consumption of the load is $P_L(t) = P_a(t) - P_{pv}(t)$, where $P_a$ is the real demand of the load and $P_{pv}$ is the power supplied to it by the PV. 
Thus, when $P_L(t) \geq 0$, the demand will be supplied by the grid or the storage device, and when $P_L(t) < 0$, the excess will flow to the battery until $E_{\max}$, the storage capacity, is reached. It is assumed that all excess power that cannot be stored in the battery will be curtailed. 

The decision variable is the power procured from the grid $P_g(t)$ which is characterized by a fuel consumption function $f(P_g(t)) \in \mathbb{R}_{\geq 0}$ which indicates the fuel needed to generate the power consumed at time $t \geq 0$. More generally, this cost function may represent a general cost function with various objectives, such as carbon emission influenced by power generation. Following \citet{Levron2010Optimal} we assume this function is twice differentiable and strictly convex, i.e. $f''(P_g(t)) > 0$. Thus, we aim to minimize
\begin{equation}
    F := \int_0^T f(P_g(t)) \mathrm{d}t,
\end{equation}
which is the total cost of generated power.

\subsection{Case-Study I: Ideal Storage}

In this setting, we assume an ideal storage device, i.e., no power is lost when charging and discharging. Accordingly, the rate of the storage charge and discharge process is given by $P_s(t) = P_g(t) - P_L(t)$ where $P_s: \mathbb{R}_{\geq 0} \to \mathbb{R}$ maps time steps to the power that flows into the storage device. 



This gives rise to the following optimization problem:
\begin{equation}
\begin{aligned}
    \underset{\{P_{g}(\cdot)\}}{\text{minimize }} & F,\\
    \text{subject to } E_g(t)&= \int _0^t P_{g}(\tau) \mathrm{d}\tau,\\
     E_L(t) &\leq E_g(t) \leq E_L(t)+E_{\max}\\
     \text{for } 0& \leq t\leq T,\\
     E_g(0) &= E_L(0),\\
     E_g(T) &= E_L(T).
\end{aligned}
\label{eq:optimization-case-1}
\end{equation}

\subsection{Solution Approaches}
Under the assumptions specified above, finding an optimal storage strategy can be considered as a Shortest-Path (SP) problem with deterministic action outcomes. 
To apply SP, we follow \cite{Levron2010Optimal} which states that the optimal generated energy $E_{g}(t)$ follows a shortest path between the bounds $E_{L}(t)$ and $E_{L}(t) + E_{\max}$. 
Along the path, the generated power satisfies the load requirement $E_{L}(t)$ and complies with the battery’s capacity $E_{\max}$. 

These assumptions set the bound of the search space within which we apply a Dijkstra search to find a shortest path between two constrained points; at times $t=0$ and $t=T$, the load is satisfied by generation, i.e. $E_g[0]=E_L[0]$ and $E_g[T]=E_L[T]=0$ 



The search graph $\mathcal{G}=(\mathcal{V}, \mathcal{E})$ has vertices $\mathcal{V} = \{v_i\}_{i=0}^N$ that represent the state of generation at time $i$ 
$$\mathcal{V} = \left \{ v_i = E_g(t_i) \mid E_L(t_i) \leq E_g(t_i) \leq E_L(t_i) + E_{\max} \right \}$$
The space between the two constraints imposes the valid transitions between energy levels of the battery
and directed edges $\mathcal{E} = {\{e_i\}_{i=0}^N}$ represent the acquisition of power from the generator.
$$\mathcal{E} = \{ e_i = (v_i = E(t_i), v_{i+1} = E(t_{i+1}))$$
$$
\mid \Delta E = E(t_{i+1}) - E(t_{i}) \in [0, E_{\max}] \}$$

Since we assume that there is no loss, the value of each vertex induces the state of charge (SOC) at time $t$ and each  edge induces the corresponding charge/discharge actions. 
The weight of each edge $e_i$ is represented by $\mathcal{W} = \{w_i \mid w_i \in \mathbb{R}\}$ and is determined by the cost of generation at its end (target) vertex. This means that for each edge $e_i$ that connects the vertices $(v_i, v_{i+1})$, the weight of the edge  is  $w_i = f(P_{g,i+1})$ where $P_{g,i+1}$ is the generated power at the state represented by the vertex $v_{i+1}$. 




The space of feasible energy levels at each discrete time step is continuous. Since search in continuous space will exceed computational complexity practical limits, we transfer the space into discrete points by grid-based uniform discretization with $N=T$ possible states at each time $t$.
Given the proposed structure, the Dijkstra search is standard. It starts at time $t=0$ and iteratively extracts edges of increasing weight until the goal condition at time $T$ is reached, and returns a lightest cost trajectory.

Importantly,
the Dijsktra-based solution relies on a function $P_L(t)$ that deterministically specifies the load at each time step within the examined time interval $[0, T]$ for the examined microgrid. 
The assumption that the load profile can be estimated with reasonable accuracy is justified for some settings and types of loads, and is addressed in forecasting problems \cite{Vercamer2016Predicting,Sandels2014Forecasting,Khan2018Approach}.

The \RL~formulation replaces the use of a deterministic load function with a generative approach. Here, the state of the environment is represented by a continuous state space $ \mathcal{S} = \{s = (E_{s}, H, P_{L}) \}$ where $E_s$ is the SOC at the current time step, $H$ is the current hour, and $P_{L}$ is current load. At each time step $t$, the agent decides which action $a \in \mathcal{A}$ to take which sets the amount of energy to charge the battery. Thus, the action space is the same as the one defined from the search problem above and is defined as $ \mathcal{A} = \{a = \Delta E, 0 \leq \Delta E \leq E_L + E_{\max}\}$ where $\Delta E \in \mathbb{R}$ is the amount of energy that should be added to the current energy level of the battery. Value bounds are dictated by the current state of charge (SOC) and the battery's capacity. 

After performing each action, the agent receives as an observation the current state 
and a scalar reward $r$
and transitions to the next state. 
The reward may be represented in different ways. 
In the simplest case, the reward function is defined by a commonly used quadratic cost function \cite{quadratic-cost-function}. The reward function is then $r = - f(P_{g})$ where $f(P_{g}) = P_{g}^{2}(t)$. 
Here, we assume that the transition function \transitionFunc~deterministically maps the current state of the battery to the prescribed SOC. Thus, after an action transfers the state $s$ into a new state $s'$, in which the agent will decide on the next action to perform.  This interaction proceeds until the agent's policy converges.



\subsection{Case-Study II: 
Lossy Storage Devices}

So far, we assumed that the amount of energy that is charged is equivalent to the energy that can be discharged. We now consider possible loss of the storage device that may occur due to internal resistance, chemical reactions, and heat generation during charging and discharging cycles.

We define $\eta: \mathbb{R} \to [0,1]$ to be a piece-wise continuous function, which is characterized by the efficiency of the storage device, and is given by
\begin{equation}
    \eta(P_s(t)) = \begin{cases}
                        \eta_{\text{ch}, t} (P_s(t)) \cdot \eta_{\text{decay}}, & \text{if $P_s(t) > 0$}\\
                        \eta_{\text{decay}}, & \text{if $P_s(t) = 0$}\\
                        \eta_{\text{dis}, t}^{-1} (P_s(t)) \cdot \eta_{\text{decay}}, & \text{if $P_s(t) < 0$}
                    \end{cases}.
\end{equation}
Therefore, the differential equation that describes the stored energy of the battery over time is given by:
\begin{equation}
    \frac{\mathrm{d}}{\mathrm{d}t} E_s(t) = \eta(P_s(t))P_s(t).
\end{equation}

\subsection{Solution Approaches}
A common approach to solving optimal storage control uses the
Pontryagin’s Minimum Principle \cite{Zargari2019Optimal}, which is widely used for controlling complex systems in which the objective is to optimize performance while accounting for this form of physical dynamics.

To apply the Pontryagin's minimum principle to our optimization problem, there is a need to eliminate the inequality constraint on the storage energy as presented in \ref{eq:optimization-case-1}. We do this by representing the constraint with cost function $c(E_s(t))$. This leads to the following formulation:
\begin{equation}
\begin{aligned}
    \underset{\{P_s(\cdot)\}}{\text{minimize }} & \int_{0}^{T} f(P_s(t)) + c \left(E_s(t) \right)) \mathrm{d}t,\\
    \text{s.t. } \frac{\mathrm{d}}{\mathrm{d}t} E_s(t) & = \eta(P_s(t))P_s(t), \\
     E_s(0) & = 0,\\
     E_s(T) & = 0, 
\end{aligned}
\end{equation}
where the function $c: \mathbb{R} \to [0,\infty)$ is defined as
\begin{equation}
    c \left( x \right) := \begin{cases}
                  \frac{\mathcal{Q}}{2E_{\max}} (x - E_{\max})^2, & \text{for $x > E_{\max}$}\\
                  0, & \text{for $0 < x < E_{\max}$}\\
                  \frac{\mathcal{Q}}{2E_{\max}} x^2, & \text{for $x < 0$}\\
                \end{cases},
\end{equation}
with $\mathcal{Q} > 0$.
By setting $\mathcal{Q}$ to be large we implicitly guarantee that the power $P_s(t)$ that flows into the battery is bounded by $[0, E_{\max}]$, else costs are extremely high. 
Consequently, the explicit analytical solution is described by the following equations:
\begin{enumerate}\label{eq:pontryagin-solution}
    \item $\frac{\mathrm{d}}{\mathrm{d}t} \hat{E}_s(t) = \eta \hat{P}_s(t)$, where $\hat{E}_s(0) = \hat{E}_s(T) = 0$
    \item $\frac{\mathrm{d}}{\mathrm{d}t} \hat{\lambda} = c \left(\hat{E}_s(t) \right)$
\end{enumerate}
where $\hat{E}_s$ is the optimal solution, $\hat{\lambda}$ is a Lagrange multiplier associated with the state constraints in the control problem, and the function $\hat{P}_s(t)$ is given by
\begin{equation}
    \hat{P}_s = \begin{cases}
        \eta_{\text{ch}} \hat{\lambda} - P_L(t), & \text{ for }P_{L} < \eta_{\text{ch}} \hat{\lambda}\\
        \eta_{\text{dis}}^{-1} \hat{\lambda} - P_L(t), & \text{ for } P_{L} > \eta_{\text{dis}}^{-1} \hat{\lambda}\\
        0, & \text{otherwise}
    \end{cases} 
\end{equation}
The \RL~formulation here is almost identical to the one presented for the ideal storage case. The only difference is expressed in the formulation of the MDP's transition function $\transitionFunc$. Instead of a direct mapping between charge and discharge actions to prescribed SOC, there is a need to account for the transmission loss. Accordingly, action $a = \Delta E$ is a legal transition from the state $s = E_s$ to $s' = (E_s + \Delta E) \cdot \eta$  if the transition respects the battery's capacity, i.e., $E_s + \Delta E < E_{max}$. To account for the loss, if $\Delta E \neq 0$ the decay is $\eta = \eta_{\text{dis}}\cdot \eta_{\text{decay}}$. Otherwise, when no charge or discharge is applied, $\eta = \eta_{\text{decay}}$.

We note that a key difference between the formulations is that while the classical solution approach needs to be adapted to account for the constraints, using \RL~alleviates the need to explicitly account for the imposed constraints with the solution approach, but rather accounts for them only within the definition of the transition function.

\subsection{Case-Study III: Accounting for Transmission Loss} 
Having accounted for the loss of the storage device, we now consider transmission loss, i.e., the power lost when it transmissions from the generator to the load or storage. While the battery charging and discharging losses were modeled as linear, we consider here non-linear losses that are induced by the transmission lines. We assume short transmission lines (length $\leq 80$ km) are used. Thus losses are due to the power lost for heat, represented by a resistor, and power losses due to the line inductance. 

We use an analytical model that we developed to analyze the losses over the transmission lines for a system that consists of a synchronous generator and a non-linear load that includes a photovoltaic unit, a storage device with capacity $E_{\max}$ and a load that consumes active power $P_L$. We assume that the elements encapsulated by the non-linear load are close to each other, so the transmission between these components is lossless. 

We aim to compute the active power $P_g$ generated before transmission loss, which yields the required power $P$ while considering the given quantities $P, V_g, R \text{ and } L$. A widely used assumption is that $P$ is low enough so that there is no voltage loss, i.e., $|V| \approx |V_g|$. 
Based on this, we can assume that the current amplitude $|I|$ can be written as $|I| = \frac{P}{|V|} \approx \frac{P}{V_g}$. Using power conservation considerations results in 
\begin{equation}
    P_g \approx P + |I|^2 R \approx P + \frac{R}{|V_g|^2} P^2.
\end{equation}

Such a quadratic loss is a well-known result in the literature (e.g., \cite{Hobbs2008Improved}). This relation encapsulates the transmission losses which will be incorporated into the cost function in the form of additional quadratic term.

This leads to the following optimization problem:
\begin{equation}
    \begin{gathered}
  \min \;\;\;\;\int\limits_0^T {f\left( {P(t)+P_L(t) + \alpha {(P(t)+P_L(t))^2}} \right)dt}  \hfill \\
  {\text{s}}{\text{.t}}{\text{.}}\;\;\;\;\;\;\frac{{dE}}{{dt}} = \eta \left( {E\left( t \right)} \right) \cdot P\left( t \right) \hfill \\
  \;\;\;\;\;\;\;\;\;0 \leqslant E\left( t \right) \leqslant {E_{\max }} \hfill \\
  \;\;\;\;\;\;\;\;\;E\left( 0 \right) = 0 \hfill \\
\end{gathered}
\end{equation}

\subsection{Solution Approaches}
In this setting, we cannot use Pontryagin’s minimum principle or shortest path algorithm used above since they can’t handle quadratic and non-linear power losses. This leads to the need to adopt a dynamic programming approach in which the time steps are discretized.

ne the time resolution $\Delta = T/N$, $i=0,...,N$ and $t=i\Delta$. Accordingly, the energy over a single interval is $E_i=E_s(i\Delta)$, and the power values are $P_i=P(i\Delta), P_{L,i}=P_L(i\Delta)$ and $P_{g,i} = P_{g}(i\Delta)$. The discrete version of the same problem is given by: 
\begin{equation}
\begin{aligned}
    \underset{\{P_{g,i}\}}{\text{minimize }} & \Delta \sum_{i=1}^N f(P_i +  P_{L,i} +\alpha (P_{i}+ P_{L,i})^2), \\
    \text{s.t. } \frac{E_i - E_{i-1}}{\Delta} & = \eta(E_i)P_i, \text{ for } i = 1,...,N\\
    0 & \leq E_i \leq E_{\max}, \text{ for } i = 1,...,N\\
    E_0 & = 0.
\end{aligned}
\end{equation}
where $\eta$ represents the aggregated energy loss.

For the RL solution, we use the model of the transmission loss over transmission lines, thus the transition from state $s = E_s$ upon taking an action $a = \Delta E$ results in a new state $s' = (E_s + \Delta E \cdot \eta_{\text{tr}}) \cdot \eta$ where $\eta_{\text{tr}}$ represents the transmission line losses, and $\eta$ is the same as in the previous case-study.

\section{Empricial Evaluation}
The objective of our empirical evaluation is to examine the quality of storage control policies achieved by generative model-free \RL~methods and their performance loss compared to classical solution methods.

\subsection{Dataset}
Our evaluation environment is a pyhton implementation of the micro-grid depicted in Figure~1, with a generator, a PV unit, a component representing the  load, and a storage unit (our complete code base and datasets can be found in the supplementary materials). 
Our dataset comprises 24-hour episodes, with half-hour intervals. At each time step, the observation includes the current load and PV production. We assume that the power produced by the PV is used to satisfy the load. This means that the only power that can be stored is surplus production from the PV or power acquired from the generator. The decision (action) is therefore how much power to generate (or buy from the generator) at each time step. Since the load must be satisfied, the generation actions together with the current SOC implicitly induce the charge and discharge action. The cost function of the generated power is of the form $f(x) = x^2$. 

\begin{figure}[htbp]
    \centering
    \includegraphics[width=0.7\linewidth]{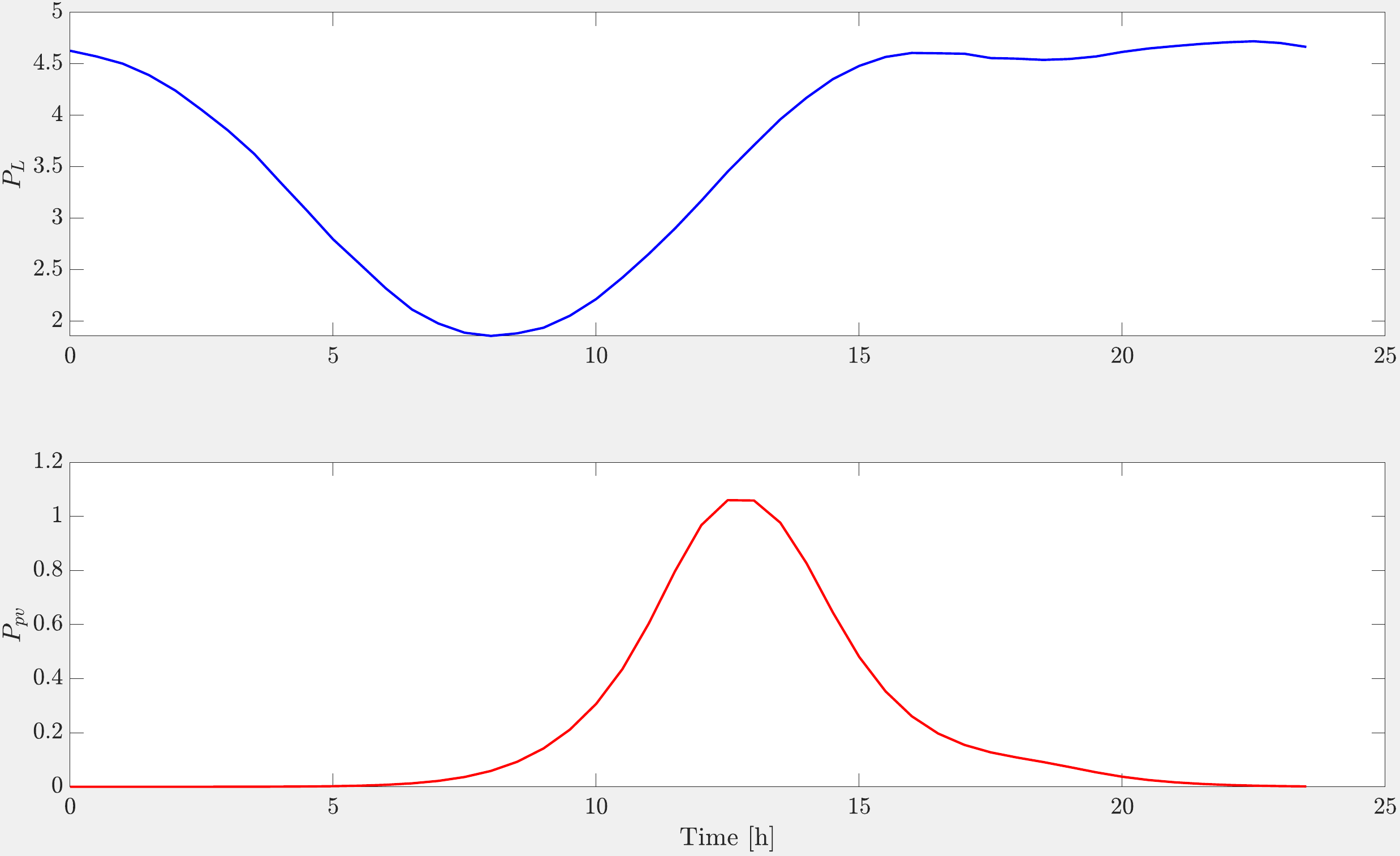}
    \caption{Average load [top] and PV generation [bottom] values over a 24-hour interval.}
    \label{fig:load-and-pv}
\end{figure}

We examined $100K$ episodes, each corresponding to a single day with 48 time steps. The dataset of the load and PV profiles was taken from a simulator proposed in \cite{Chowdhury2021Optimal}. 
We extracted from the dataset (with its 10K episodes) a test set of 100 episodes, while the rest served as the training set.
Figure \ref{fig:load-and-pv} shows the average load and pv production over a 24-hour episode for the test set. 

We consider the three case-studies described above: 
\begin{itemize}
    \item Case-Study I: Ideal Storage (IS)
    \item Case-Study II: Lossy Storage (LS)
    \item Case-Study III: Lossy Transmission (LT)
\end{itemize}

For each case-study, we implemented the models for storage and transmission described for each case-study using a Python-based implementation that we developed to simulate the different micro-grids (the complete code base and data set can be found in our supplementary materials). 

Notably, the models we developed for the transmission and storage served a different purpose for the classical and \RL~approaches. For the classical approaches, the models were used to compute the solution. For RL evaluation, the models were integrated into our simulator to set the transition function and facilitate data generation. 
\begin{figure*}[htbp]\label{fig:results-over-one-day-all-cases}
     \centering
     \begin{subfigure}[b]{0.3\textwidth}
         \centering
         \includegraphics[width=\textwidth]{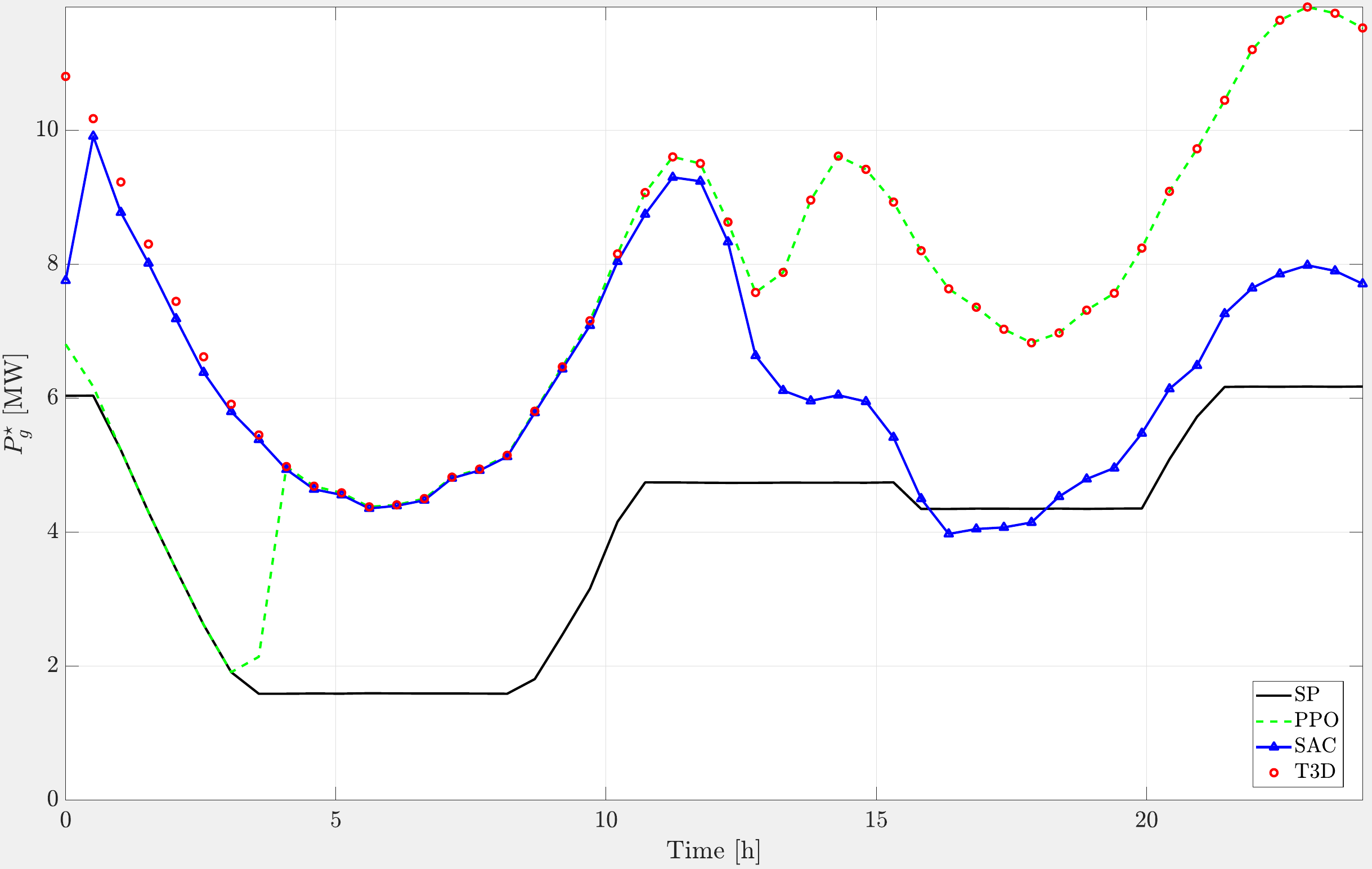}
         \caption{IS}
         \label{fig:case_1}
     \end{subfigure}
     \hfill
     \begin{subfigure}[b]{0.3\textwidth}
         \centering
         \includegraphics[width=\textwidth]{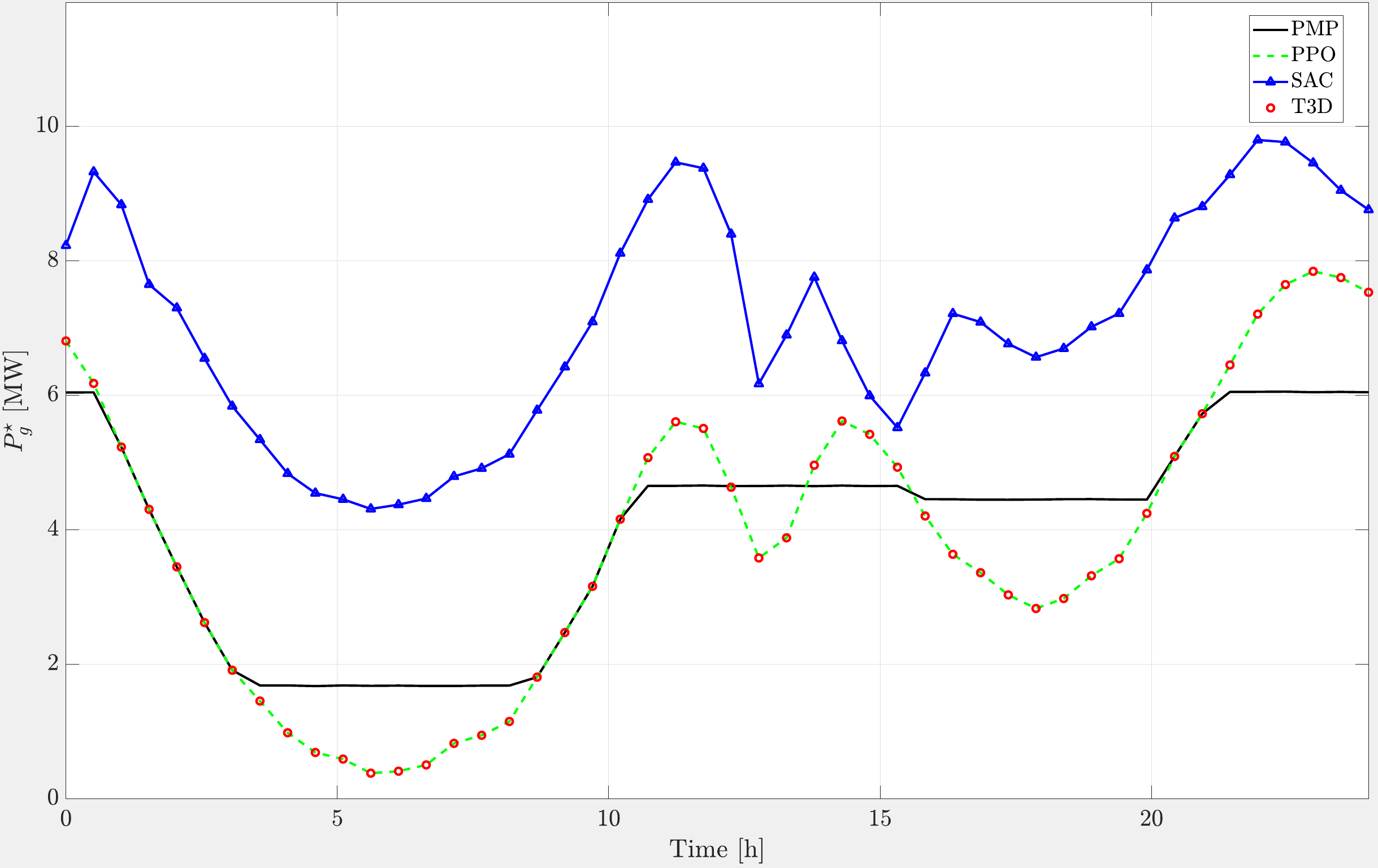}

         \caption{LS}
         \label{fig:case_2}
     \end{subfigure}
     \hfill
     \begin{subfigure}[b]{0.3\textwidth}
         \centering
         \includegraphics[width=\textwidth]{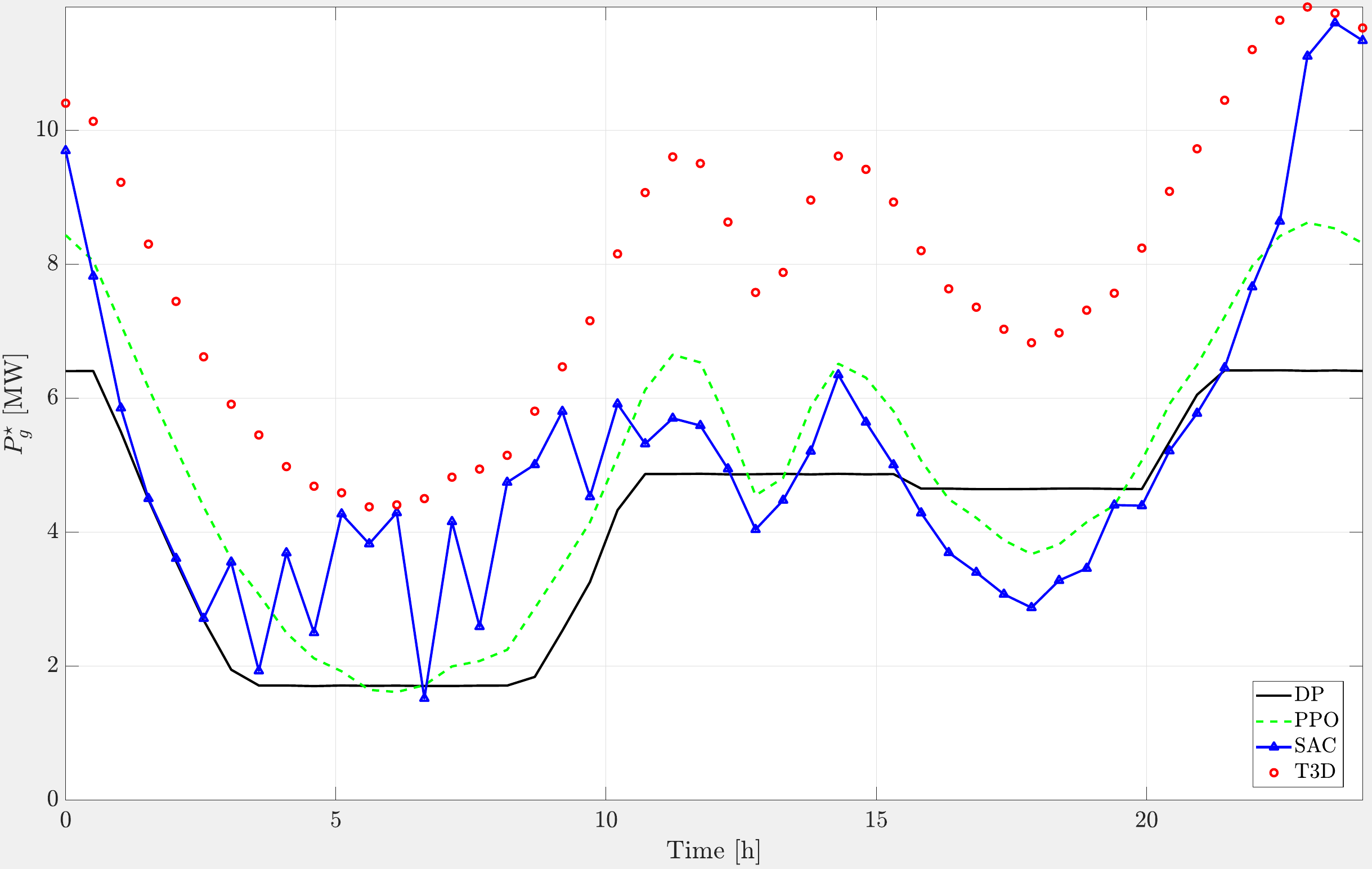}

         \caption{LT}
         \label{fig:case_3}
     \end{subfigure}
        \caption{Single day generation policy for the classical and \RL~approaches for the three study-cases.}
\end{figure*}

\subsection{Setup}
To find storage control policies for each of the case-studies, we implemented the corresponding analytical method described above in MATLAB and three model-free \RL~methods in Python. 
The solution is a policy specifying the generation (and induced charge/discharge) actions to perform over the 48 discrete time steps over a 24-hour interval. For the classical method, a solution is found for each episode in the test set for the three settings (IS, LS and LT).  
The \RL~methods were trained on the training set, and the resulting generalized policy was examined on the test set.

In this work, we focus on different model-free approaches that do not explicitly model the dynamics of the environment but rather directly optimize the policy. We use these methods since they are the most broadly used in general and for storage control in particular. This is mostly due to their general applicability and ease of adaptation to novel domains. Nevertheless, developing model-based approaches is an important and promising avenue for future investigation. The \RL~methods that we examined were Soft-Actor Critic (SAC) \cite{haarnoja2018soft}, PPO \cite{schulman2017proximal}, and TD-3 \cite{fujimoto2018addressing}. We used the implementation by Stable-Baselines \cite{stable-baselines3}.

  For each method, we conducted a series of preliminary tests to tune the key hyperparameters, such as learning rates, discount factors, and network architectures, based on the performance metrics specific to our environment (the full list of hyperparameters and their values can be found in the supplementary material). For this we used a validation set that included 10 episodes applied after every 10k iterations. After tuning, we employed the same hyperparameters of each method across the three case studies to ensure a fair comparison.
 These hyperparameters were then fixed for the final evaluation phase to avoid any bias introduced by varying the parameters during training. 

\subsection{Results}
To examine the performance of the RL methods, we compare the expected utility of the \RL~policy generated by each method for the three case studies to the expected utility of the policy generated by the classical methods (best result in bold). 
Table~\ref{table:mse-rl-perfomance} presents for each case study, the normalized MSE between each \RL~algorithm and the classical solution over 100 episodes.
 An interesting observation is that results show that the worst performance is achieved for the simplest IS case, in which there is no loss. In contrast, for the LS setting, the error is negligible for the  PPO and TD3 methods. Similarly, for LT PPO achieves the best performance. 

\begin{table}
    \centering
    \begin{tabular}{|c|c|c|c|}
    \hline
          & IS & LS & LT  \\
          \hline
        SAC & {\bf 0.59} & 1.00   & 0.26  \\
        \hline
        PPO & 0.83 & {\bf 0.027}  & {\bf 0.19}  \\
        \hline
        TD3 & 1.00 & {\bf 0.027} & 1.00 \\
        \hline
    \end{tabular}
    \caption{Normilized MSE calculation for all use-cases.}
    \label{table:mse-rl-perfomance}
\end{table}

Figure 3 offers a deeper investigation of the results achieved for a single day (additional days, as well as the average results achieved per time-step, can be found in the supplementary material). The $x$ axis represents the half-hour intervals, and the $y$ axis represents the cost that is incurred for acquiring power from the generator. The trend depicted in these results aligns with the results depicted above. Accordingly, none of the \RL~methods managed to learn an effective policy for the IS case, when compared to the optimal path found by the state space search. In contrast, both PPO and TD3 generated a policy that is similar to that achieved by the classical solution. An interesting observation is that the policy of both methods is similar for the LS case, although the two policy-search methods differ in that T3D uses a deterministic rather than stochastic policy. Another interesting observation is that for the LT case, the policy produced by PPO is not only better in terms of its performance but also in terms of its smoothness. While this is not reflected in the cost function, an interesting and important avenue for future work is to try and find ways to penalize methods for abrupt changes in the charge and discharge values.

A key insight from our investigation is that using the complex dynamics models within a simulator can, in some cases, replace the need to apply highly complex methods for solving the settings that require accounting for loss in the storage and transmission steps. However, it is worth investigating whether there are additional ways of using these models to improve, for example, the way data is collected, the priors in the learning process, and the way by which the learning process is regulated. 

\section{Conclusion}
In recent years, energy storage devices have emerged as crucial components in power systems, and are a critical component for integrating renewable energy sources and maintaining grid stability. Moreover, the escalating dimensionality of storage systems used in power systems requires the development of advanced algorithms for handling the associated control problems. To address this challenge, we aim in this paper to better understand the tradeoffs between traditional and Reinforcement Learning approaches for storage management. More specifically, we wish to better understand the performance loss incurred when the physical model of the storage component is not known, and a data-driven approach is used instead. Our results show several interesting trends: First and foremost, it is evident from the results that if perfect knowledge about the physical model exists, one can achieve far superior results to those achieved by statistical learning. However, when using large amounts of data for the training set, the results may be comparable in certain cases, but not in all of them. Moreover, while traditional solutions always work for every load profile and renewable-energy generation profile, model-free solutions require re-training for every new scenario, and cannot be immediately transferred, for instance, from one household to another. This highlights the importance of using model based approaches, or at-least incorporating some information on the system dynamics into the training process, which allows the model to better generalize, and to utilize the knowledge acquired in one problem to another, even if their underlining statistics are not exactly identical.

\section*{Acknowledgments}
\small 
This research was carried out within the framework of the Grand Technion Energy 
Program (GTEP)

\bibliography{ref}
\end{document}